\documentclass[letterpaper, 10 pt, conference]{ieeeconf}  % Comment this line out if you need a4paper

\IEEEoverridecommandlockouts                              % This command is only needed if
\makeatletter
\let\NAT@parse\undefined
\makeatother
\usepackage[sort&compress, numbers]{natbib}

\usepackage[pdftex]{graphicx}
\usepackage{amsmath}
\usepackage{amssymb}  % assumes amsmath package installed
\usepackage{subfig}
\usepackage{multirow}
\usepackage{array,booktabs}
\usepackage{diagbox}
\usepackage{balance}
\usepackage[ruled]{algorithm}
\usepackage{algpseudocode}
\usepackage{float}
\usepackage{tabularx}
\usepackage{bm}
\usepackage{pifont}
\newcommand{\cmark}{\ding{51}}
\newcommand{\xmark}{\ding{55}}

 \usepackage[final]{hyperref}
 \hypersetup{
     colorlinks=true,
     linkcolor=blue,
     filecolor=magenta,
     urlcolor=blue,
     citecolor=black
 }
\usepackage[labelfont=bf, font=small]{caption}
\usepackage{./rpm_packages/rpm_SIunits}
\usepackage{./rpm_packages/rpm_acronyms}
\usepackage{./rpm_packages/rpm_math}
\usepackage{./rpm_packages/rpm_misc}

\usepackage{soul,color}
\usepackage{lipsum}

\usepackage{xcolor}

\newcommand{\bl}[1]{{\textcolor{blue}{#1}}}

\usepackage{tikz} % \checkmark

\title{\LARGE \bf Co-RaL: Complementary Radar-Leg Odometry with \\4-DoF Optimization and Rolling Contact}

\author{Sangwoo Jung${}^{1}$, Wooseong Yang${}^{1}$,  and Ayoung Kim${}^{1*}$
\thanks{$^\dagger$This work was supported by the MOTIE (1415187329) and MSIT (No.2022-0-00480).}
\thanks{$^{1}$S. Jung, W. Yang, and A. Kim are with the Department of Mechanical Engineering, SNU, Seoul, S. Korea {\tt\small [dan0130, yellowish, ayoungk]@snu.ac.kr}}%
}

\makeatletter
    \def\tagform@#1{\maketag@@@{\normalsize(#1)\@@italiccorr}}
\makeatother

\begin{document}

%\onecolumn
\maketitle
\thispagestyle{empty}
\pagestyle{empty}

\begin{abstract}

Robust and accurate localization in challenging environments is becoming crucial for SLAM.
In this paper, we propose a unique sensor configuration for precise and robust odometry by integrating chip radar and a legged robot.
Specifically, we introduce a tightly coupled radar-leg odometry algorithm for complementary drift correction.
Adopting the 4-DoF optimization and decoupled RANSAC to mmWave chip radar significantly enhances radar odometry beyond the existing method, especially z-directional even when using a single radar.
For the leg odometry, we employ rolling contact modeling-aided forward kinematics, accommodating scenarios with the potential possibility of contact drift and radar failure.
We evaluate our method by comparing it with other chip radar odometry algorithms using real-world datasets with diverse environments while the datasets will be released for the robotics community. \bl{https://github.com/SangwooJung98/Co-RaL-Dataset}
% We will release our self-collected dataset for our community.

\end{abstract}
\section{Introduction}
\label{sec:intro}

% Radar and legged robots have recently emerged in addressing extreme environments where traditional vision sensors and wheeled \ac{UGV} face significant challenges. 
% Robust and precise odometry estimation in extreme environments is essential for the practical application of mobile robots in the real world. 
% This leads to the \ac{SLAM} in the extreme environment to be a concentrated topic~\cite{burnett2022we, giubilato2022challenges, hong2020radarslam, sheeny2021radiate, zhang2023smart} which requires both robustness in perception and agility in mobility. 
% However, traditional vision sensors such as camera and \ac{LiDAR} are prone to visual degradation and feature degeneration, leading to sensor failure. 
% Traditional wheel \ac{UGV} may operate well in smooth and flat ground; environments with non-standard terrains, such as stairs or mountain paths, are left as obstacles. 

% Acquiring precise and robust odometry in extreme environments, one of the recent topics on \ac{SLAM}~\cite{burnett2022we, giubilato2022challenges, hong2020radarslam, sheeny2021radiate, zhang2023smart}, requires both robustness in perception and agility in mobility. 
% Consequently, radar and legged robots are one of the rising sensors and \ac{UGV} for unstructured and extreme environments. 

Attaining precise and robust odometry in extreme environments is receiving a prominent focus~\cite{burnett2022we, giubilato2022challenges, hong2020radarslam, sheeny2021radiate, zhang2023smart}. 
Navigating such environments demands robust perception and agile mobility.
Providing such dexterous mobility, legged robots have emerged recently due to their compatibility in unstructured environments. For most legged robots, common practice is to equip the legged robot with cameras and \ac{LiDAR}s for their extrovert perception.
However, these perceptual sensors are susceptible to visual degradation and feature degeneration, leading to sensor potential failures. 
Being a robust perceptual sensor in extreme environments, radar can be a promising solution but mainly applied to \ac{UGV}s and drones, not to the legged robots.
This paper presents a tightly coupled radar-leg odometry, validating the enhanced odometry performance from their complementary nature.
%Conventional wheeled \ac{UGV} operate well in smooth and flat ground; environments with non-standard terrains, such as stairs or mountain paths, are left as challenges. 
%In this manner, radar and legged robots have emerged as promising alternatives for unstructured and extreme environments. 
% djlee
% Recent research in simultaneous localization and mapping \ac{SLAM} has placed significant emphasis on achieving precise and robust odometry, particularly in extreme environments~\cite{burnett2022we, giubilato2022challenges, hong2020radarslam, sheeny2021radiate, zhang2023smart}. Navigating such environments demands robust perception and agile mobility. However, traditional vision sensors such as cameras and \ac{LiDAR} are susceptible to visual degradation and feature degeneration, leading to potential sensor failures. Conventional wheeled \ac{UGV} perform well on smooth and flat terrain but encounter challenges in non-standard environments such as stairs or mountain paths. As a result, radar sensors and legged robots have emerged as promising alternatives for navigating unstructured and extreme terrains.

\begin{figure}[!t]
    \centering
    \includegraphics[width=\columnwidth]{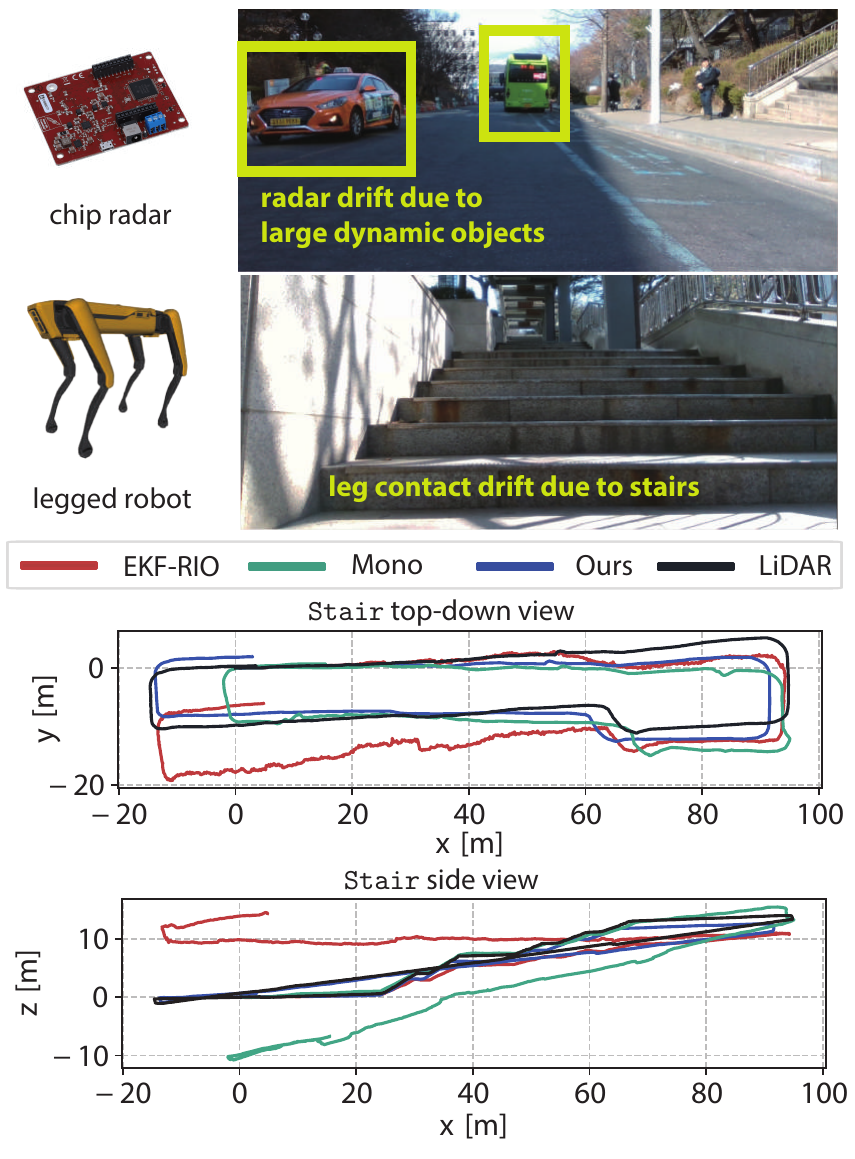}
\vspace{-7mm}
    \caption{Example of radar drift scene that includes large dynamic objects and leg contact drift scene including frequent contact impact and vibration. 
    Large dynamic objects are shown in the yellow box. 
    Both scenes are recorded using the intel realsense D435i attached to the SPOT sensor system while acquiring \texttt{Stair} sequence. Odometry of the proposed method (\texttt{Ours}) and comparison group are included in top-down and side views.}
\vspace{-5mm}
    \label{fig:fig_1}
\end{figure}

Radar stands out as a widely utilized range sensor in robotics, renowned for its robustness in extreme environmental conditions such as rain, dust, or snow. 
%From its extended wavelength compared with vision sensors such as cameras and \ac{LiDAR}, radar penetrates small particles, providing reliable information in various environments. 
Furthermore, \ac{FMCW} technique enables the radar to gather the radial velocity of points included in the point cloud, which may exploited for noise or small dynamic object reduction and ego-velocity calculation~\cite{kellner2013instantaneous}. 
However, challenges arise from the sparse point cloud and low elevation accuracy, leading to potential vertical drift in estimating odometry as \figref{fig:fig_1}. 
%To address this, some approaches involve rejecting elevation information and integrating multiple radars to acquire 3D odometry~\cite{kellner2014instantaneous, park20213d}. 
%Integration with non-exteroceptive information is essential to handle radar failures, especially on the sensor occlusion by large dynamic objects that may cause ego-velocity drift.

% djlee
% Radar stands out as a widely utilized range sensor in robotics, renowned for its robustness in extreme environmental conditions such as rain, dust, or snow. Its extended wavelength, compared to cameras and LiDAR, allows it to penetrate small particles, providing reliable information in diverse environments. Furthermore, the \ac{FMCW} technique enables the radar to gather radial velocity for each point in the point cloud, facilitating noise reduction and ego-velocity calculation~\cite{kellner2013instantaneous}. However, challenges arise from the sparse point cloud and limited accuracy in elevation, leading to vertical drift. To address this issue, some approaches involve forsaking elevation information and integrating multiple radar sensors to acquire odometry~\cite{kellner2014instantaneous, park20213d}. Moreover,  integrating non-exteroceptive odometry becomes essential to mitigate radar sensor failures, particularly when occluded by large dynamic objects, which could introduce ego-velocity drift.

Legged robots, in contrast to traditional wheeled \ac{UGV}, show superior adaptability to varied terrains, including challenging environments like fire sites, mountain trails, and stairs, due to their reduced dependence on ground structure. These robots leverage contact sensors and joint encoders to compute odometry. The estimated velocity from legged robots is not influenced by dynamic objects and offers accurate z-axis information, thanks to forward kinematics. Nonetheless, variations in ground conditions can lead to contact slip or drift, which may compromise the reliability of leg measurements. To address this, the integration of cameras and \ac{LiDAR}s has been proposed to enhance leg odometry robustness with external sensors~\cite{wisth2022vilens}. However, it's crucial to note that the advantages offered by cameras and \ac{LiDAR}s might be mitigated in extreme conditions---environments where legged robots typically excel---due to the susceptibility of these sensors to such challenging scenarios.

\begin{figure}[!t]
    \centering
    \includegraphics[width=\columnwidth]{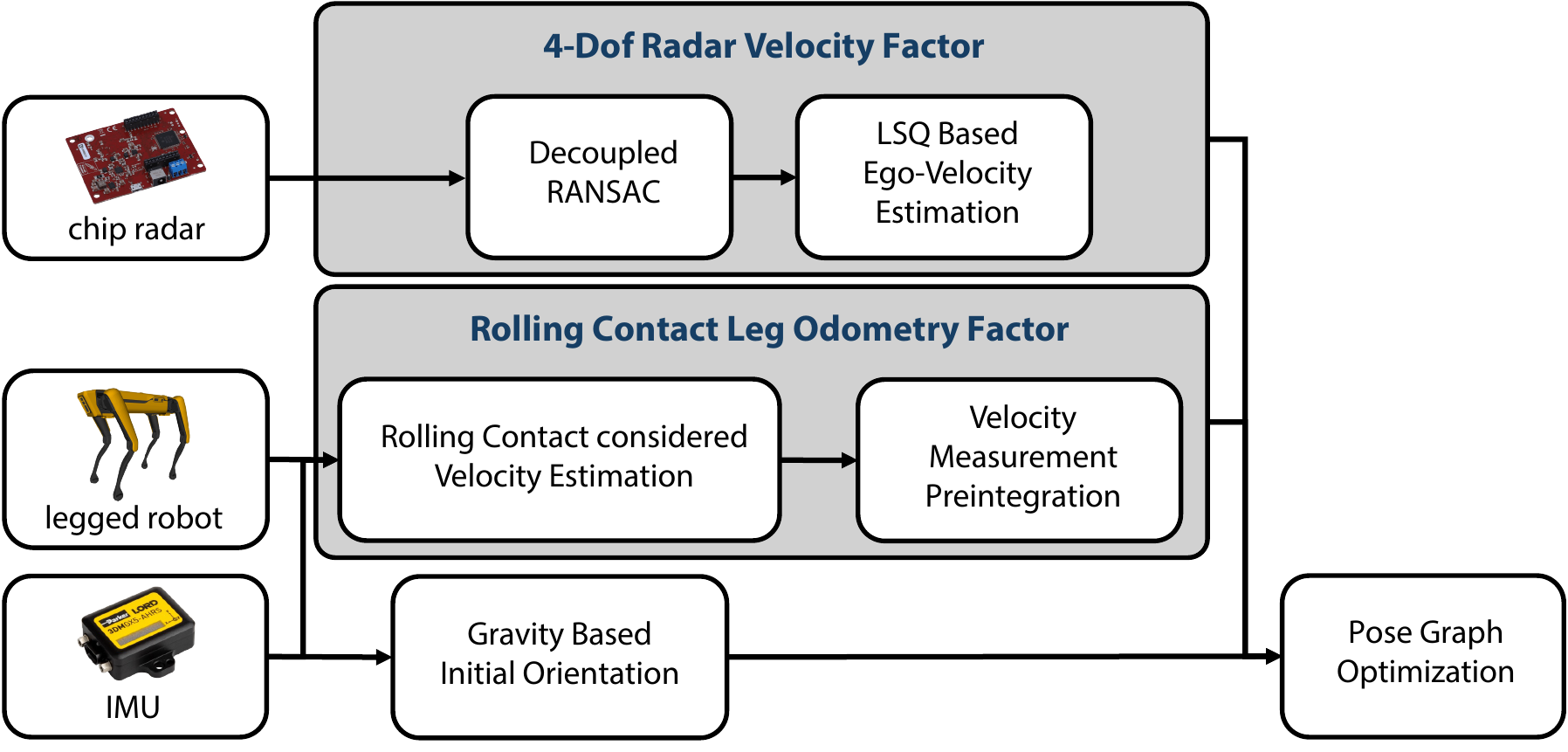}

    \caption{Pipeline of the proposed system with 4-DoF radar velocity factor and rolling contact preintegrated leg odometry factor. }
\vspace{-5mm}
    \label{fig:pipeline}
\end{figure}

To the best of our knowledge, our system is the first integration of radar and leg for ego-motion estimation. 
We introduce a tightly integrated radar-leg odometry designed to mutually correct drifts (\figref{fig:pipeline}). By transforming the challenge into a 4-\ac{DoF} problem and employing \ac{RANSAC} on two decoupled planes, we enhance radar odometry beyond the existing methods even with using a single radar. Additionally, we account for the rolling-contact modeling in leg odometry, accommodating scenarios such as stair navigation.
%In this paper, we propose two complementary factors for pose graph-based odometry: a 4-DoF optimized chip radar factor and a rolling contact considered preintegrated leg kinematic factor. 
%Two factors are tightly coupled to calculate the accurate odometry, especially in the z-axis direction. 
Our contributions can be summarized as follows:
% djlee
% 갑자기 factor라는 단어가 등장하는데, 이게 pose graph를 이용한 odometry 논문이라는 말이 abstract말고는 등장하지 않아서 그 부분을 추가하면 좋을 것 같습니다
% In this paper, we propose two complementary factors for pose graph-based odometry: a 4-DoF optimized single-chip radar factor and a rolling contact considered a preintegrated leg kinematic factor. Two sensors, the single-chip radar and contact sensor, are tightly coupled to calculate accurate odometry, especially in the z-axis direction.

\begin{itemize}
    \item \textbf{Substantial z-directional accuracy improvement} 
    We are the first to apply the 4-\ac{DoF} optimization scheme to radar, which is widely leveraged on the vision group \cite{kubelka2022gravity, qin2018vins}. Additionally, we introduce decoupled \ac{RANSAC} specifically for z-directional outlier removal. Proposed method substantially improves the radar odometry in the z-direction, even using a single-chip radar.
    
    \item \textbf{Rolling contact for preintegrated leg odometry} This paper handles the rolling contact movement of the contact surface by including the rolling motion on the contact surface to the body velocity derivation, the velocity measurement model with less bias is achieved. 
    
    \item \textbf{Thorough real-world evaluation} Using a real-world radar-leg dataset, we evaluate the proposed method by comparing it against other existing algorithms. Our method shows substantially improved accuracy and robustness, especially in the z-axis translation. We will release the dataset for this unique sensor combination to boost the research in this field.
    
\end{itemize}

\section{Related Work}
\label{sec:relatedwork}

\subsection{mmWave Chip Radar Odometry}

% A recent survey paper~\cite{harlow2023new} presents the current state of millimeter-wave (mmWave) radar applications in the field of robotics. 
% This section will mainly focus on chip radar(-inertial) odometry research. 

Using radar for ego-motion estimation has been introduced in various ways. \citeauthor{kellner2013instantaneous}~\cite{kellner2013instantaneous} proposed an instantaneous ego-motion estimation using a single-chip radar, which was later extended to multiple chip radars~\cite{kellner2014instantaneous}.
%Dynamic objects and noise are removed using \ac{RANSAC} while 2D ego-velocity of the robot is predicted based on the static objects and \ac{LSQ} optimization. 
%The same authors of \cite{kellner2013instantaneous} expanded their work on 3D by exploiting multiple chip radars.
\citeauthor{doer2020ekf}~\cite{doer2020ekf} proposed an \ac{EKF} based fusion between chip radar and \ac{IMU} while exploiting the ego velocity estimation solution of \cite{kellner2013instantaneous}. 
\citeauthor{park20213d}~\cite{park20213d} estimated 3D ego-velocity in a similar manner of \cite{kellner2014instantaneous} by exploiting two perpendicular chip radar systems while proposing a radar velocity factor for pose graph \ac{SLAM} that leverages rotation information from \ac{IMU}. 
%Though the \ac{RANSAC} and \ac{LSQ} based ego-velocity estimation has advanced the usage of the chip radar to the very next level, 
Unfortunately, even when using multiple radars, inaccurate elevation due to precision of the chip radar and the inability to resolve drift caused by large dynamic objects still remain.  

Recent works focus on leveraging the radar pointcloud at the point level to include spatial information on the resulting trajectory. 
\citeauthor{michalczyk2022tightly}~\cite{michalczyk2022tightly} proposed 
%an \ac{EKF} based fusion between chip radar and \ac{IMU} while 
exploiting stochastic cloning to match 3D points on consecutive pointclouds. 
\citeauthor{zhuang20234d} presented 4D-iRIOM~\cite{zhuang20234d} that fuses ego-velocity and scan matching. % while exploiting iterative \ac{EKF}. 
By combining sensitive but accurate scan-to-submap registration with robust ego-velocity, 4D-iRIOM~\cite{zhuang20234d} achieved accurate odometry of the \ac{UGV}. 
Leveraging ground point characteristics has been done on DRIO~\cite{chen2023drio}. 
By simultaneously estimating the radar velocity and detecting ground points, DRIO~\cite{chen2023drio} accomplished robust 3D odometry. 
Recently, \citeauthor{huang2024less}~\cite{huang2024less} exploited the \ac{RCS}-bounded filter to refine point-to-point correspondences. % and tightly coupled it with doppler velocity and \ac{IMU} measurements. 
%The deep learning-based idea for sensor fusion to handle the high-level noise of radar data has been introduced by \citeauthor{lu2023efficient}~\cite{lu2023efficient}. 
More recently, the deep learning was considered to handle the high-level noise~\cite{lu2023efficient}. 
%Various attempts were introduced to exploit the spatial information of the chip radar data; 
Yet, due to its noise and sparseness, potential failure of optimization and environmental constraints should be further resolved.

We present a single-chip radar factor that is more robust on vertical drift compared with \cite{park20213d, doer2020ekf}, leveraging elevation noise removal and 4-\ac{DoF} optimization. Compared to the existing methods, ours yields improvement in z-directional error, which mutually compensates with leg odometry.
%We filter the noise and dynamic object with first \ac{RANSAC}, and reduce incorrect elevation points with second \ac{RANSAC}. 
%Furthermore, we first adapt the 4-DoF optimization scheme to minimize the z-axis error from the inaccurate roll and pitch of the orientation. 

%%%%%%%%%%%%%%%%%%%%%%%%%%%%%%%%%%%%%%%%%%%%%%%%%%%%%%%%%%%%%%%%%%%%%%%%%%%%%%%%%%%%%%%%%%%%%%%%%%%

\subsection{Legged Robot Odometry}

Being the major sensor in a legged robot, contact sensor provides information about whether the robot's foot is contacting the ground. 
\citeauthor{hartley2018legged}~\cite{hartley2018legged} proposed forward kinematics factor and preintegrated contact factor for the graph optimization. 
%The forward kinematics factor connects the pose between the contact frame and the body frame, while the integrated contact factor combines the contact frame pose, exploiting the assumption that the contact frame is fixed on the world while the contact sensor is on. 
Later, \cite{hartley2018hybrid} expands \cite{hartley2018legged} in the manner of hybrid contact theory that the contact frame may switch while integration as at least one foot is contacting on the ground. 
Based on the contact theory, \citeauthor{hartley2020contact} proposed \cite{hartley2020contact} that included the contact kinematics theory with the invariant \ac{EKF} model to integrate it with other exteroceptive sensors. 
Though the contact frames may provide reliable landmarks to the robot based on the forward kinematics, contact drift or slip may occur movement on the contact frame, which accumulates the error on robot pose. 

% 아랫 문단은 굳이 필요한가...???
%To handle the sudden slip on contact frame, \citeauthor{kim2021legged}~\cite{kim2021legged} proposed the fixed-lag smoother state estimation that conditionally exploits kinematic measurement based on the slip rejection model. 
%STEP \cite{kim2022step} proposed a preintegrated foot velocity factor that leverages the velocity of the contact frame itself as a parameter to estimate, denying the non-slip assumption of the contact frame. 
%\citeauthor{yang2023multi}~\cite{yang2023multi} exploited multiple \ac{IMU} that are attached on every calf link of the robot to detect contact and foot slip without using any contact sensors while integrating them with joint encoders on \ac{EKF}. 

% 줄이긴 했으나 이건 slip 이야기 같아서 분량상 제외함.
%To handle the sudden slip on contact frame, \citeauthor{kim2021legged}~\cite{kim2021legged} introduced a fixed-lag smoother for state estimation utilizing kinematic measurements and a slip rejection model. Later, \cite{kim2022step} developed a preintegrated foot velocity factor, considering the contact frame's velocity, challenging the non-slip assumption. \citeauthor{yang2023multi}~\cite{yang2023multi} used multiple IMUs on robot calf links for slip and contact detection without contact sensors, integrating with joint encoders on EKF.

Calculating robot velocity based on the differential of forward kinematics from the contact frame to the robot frame is one method of exploiting the joint encoder and contact sensor as a proprioceptive odometry sensor.
\citeauthor{wisth2020preintegrated}~\cite{wisth2020preintegrated} proposed a preintegrated velocity factor that includes bias estimation to overcome contact nonlinearities, being fused with the camera to calculate odometry. 
VILENS~\cite{wisth2022vilens} extended the previous work \cite{wisth2020preintegrated} by tightly fusing camera, \ac{LiDAR}, \ac{IMU}, and preintegrated leg factor from \cite{wisth2020preintegrated} to achieve robust odometry that can be exploited on feature degenerate, low light, and even deformable ground environment. 

Inspired by VILENS~\cite{wisth2022vilens}, we conduct a preintegrated velocity factor exploiting the joint encoder and contact sensor.
Differing from them, we consider the rolling contact of the contact frame. 
Separating the portion of velocity bias occurring from the rolling contact may lower the burden of optimizing velocity bias in the optimization process. 
\section{4-DoF Radar Velocity Factor}
\label{sec:radar_factor}

% This section is divided into three subsections. 
% \ref{DR} includes double \ac{RANSAC} for outlier rejection, remaining only static and z-axis accurate points. 
% \ref{LSQ} performs least square optimization to calculate the 3D ego-velocity of the robot, and \ref{RVF} introduces the 4-DoF radar velocity factor that may exploited for factor graph. 
% The pipeline of the 4-DoF radar velocity factor is included in \figref{fig:radar_factor}.

% This section is divided into two subsections. 
% \ref{DR} includes double \ac{RANSAC} to remove the dynamic objects and inaccurate points on the z-axis. After the double \ac{RANSAC} process, \ac{LSQ} optimization proceeds to calculate the 3D ego-velocity of the robot. \ref{RVF} introduces the 4-DoF radar velocity factor that may exploited for factor graphs. 

\subsection{Decoupled RANSAC for Outlier Rejection} \label{DR}

Pointcloud from chip radar provides the radial velocity of each point, but a high level of noise and inaccurate elevation information problems exist. In this section, we introduce our two-staged outlier removal. 
%In addition to the existing xy-plane \ac{RANSAC} to filter the noise and dynamic object, we further remove xz-plane noise to reduce incorrect elevation points with second \ac{RANSAC}.

\subsubsection{$xy$ Plane Outlier Removal} \label{xy_ransac}

% First \ac{RANSAC} process is the same as \cite{kellner2013instantaneous, kellner2014instantaneous, park20213d}, removing small dynamic objects and noise from the pointcloud. 
% Position and radial velocity of every point are projected on XY plane, remaining the position $(x, y)$ and velocity vector $v_{xy}$. 
% After the projection, the point position is changed into a polar coordinate system, changing into $(r_{xy}, \theta_{xy})$.
% If a point is generated from a static object, 2D radial velocity $v_{xy}$ should be the same as the robot velocity of point direction $\theta$, leading to the following equation. 
% \begin{eqnarray}
%     \left[ 
%     \begin{array}{c}
%     v_{xy, 1} \\ \vdots \\ v_{xy, n} 
%     \end{array}
%     \right]
%     = 
%     \left[
%     \begin{array}{cc}
%     cos\theta_{xy, 1} & sin\theta_{xy, 1} \\
%     \vdots & \vdots  \\
%     cos\theta_{xy, n} & sin\theta_{xy, n} 
%     \end{array}
%     \right]
%     \left[
%     \begin{array}{c}
%     v_x \\ v_y
%     \end{array}
%     \right]
% \end{eqnarray}

% For static points, $\theta_{xy}$ and $v_{xy}$ of every point should follow a sinusoidal curve~\cite{kellner2014instantaneous}. 
% Based on the \ac{RANSAC} sinusoidal curve fitting, the 2D velocity of robot $(v_x, v_y)$ and points from the static object can be extracted.

Similar to \cite{kellner2013instantaneous, kellner2014instantaneous, park20213d}, the initial \ac{RANSAC} process removes small dynamic objects and noise from the pointcloud. 
After the $xy$ plane projection, the point position is changed into a polar coordinate system, changing into $(r_{xy}, \theta_{xy})$.
If a point is generated from a static object, 2D radial velocity $v_{xy}$ should be the same as the robot velocity of point direction $\theta_{xy}$, which follows a sinusoidal curve~\cite{kellner2014instantaneous}. 
Based on the sinusoidal curve fitting, the 2D velocity of robot $(v_x, v_y)$ and static points are extracted. 

\subsubsection{z-directional Outlier Removal} \label{xz_ransac}

The extracted inlier pointcloud data with noise and dynamic objects removed may include the remaining inaccuracy in the z-axis position. 
    
The follow-up \ac{RANSAC} starts with projecting these filtered inlier points to the $xz$ plane. 
Every position information of projected points $(x, z)$ is changed into a polar coordinate $(r_{xz}, \theta_{xz})$. 
For that point whose z-axis information is accurate, $\theta_{xz}$ and $r_{xz}$ should follow the sinusoidal curve.

The x-axis velocity $v_x$ calculated from the previous phase is reliable since it only exploits the $v_{xy}$ of the static points. 
Therefore, fixing the $v_x$ may enhance the performance of \ac{RANSAC} fitting to remove the points whose elevation is inaccurate. 
After removing outliers, \ac{LSQ} optimization computes the 3D ego-velocity as $\hat{\textbf{v}} = {\left[ {\hat{v}}_x ~ {\hat{v}}_y ~ {\hat{v}}_z  \right]}^\top$.

% \subsection{3D Ego-velocity by Least Square} \label{LSQ}

% The inlier points extracted from the \ref{xz_ransac} are generated from static objects while their z-axis accuracy is reliable, leading to the following equation. 
% \begin{eqnarray}
%     \left[ 
%     \begin{array}{c}
%     v_{r, 1} \\ \vdots \\ v_{r, n} 
%     \end{array}
%     \right]
%     = 
%     \left[
%     \begin{array}{ccc}
%     \frac{x_1}{r_1} & \frac{y_1}{r_1} & \frac{z_1}{r_1} \\
%      & \vdots &  \\
%     \frac{x_n}{r_n} & \frac{y_n}{r_n} & \frac{z_n}{r_n}
%     \end{array}
%     \right]
%     \left[
%     \begin{array}{c}
%     v_x \\ v_y \\ v_z
%     \end{array}
%     \right]
% \end{eqnarray}
% \ac{LSQ} optimization is performed to compute the 3D ego-velocity as $\hat{\textbf{v}} = {\left[ {\hat{v}}_x ~ {\hat{v}}_y ~ {\hat{v}}_z  \right]}^\top$.    

\subsection{4-DoF Radar Velocity Factor} \label{RVF}

Visual and \ac{LiDAR} \ac{SLAM} works as \cite{qin2018vins, kubelka2022gravity} has exploited the characteristic that the direction of gravity vector is constant to optimize only $x$, $y$, $z$, and yaw to enhance the optimization accuracy, especially on the z-axis. 
In this work, we decouple the orientation state of the 6-DoF radar factor from \cite{park20213d} into roll, pitch, and yaw. 
After decoupling, we fix the roll and pitch values as the initial measurements of \ac{IMU} to prevent inaccurate optimization. 
Derivation in this paper only includes the residual function and Jacobian matrix to highlight the parts different from \cite{park20213d}. 

Residual error included in \cite{park20213d} are as follows:
\begin{eqnarray}
    \textbf{r}_{\Delta \textbf{p}_{ij}} &=& \textbf{R}^{\top}_{i} \left( \textbf{p}_j - \textbf{p}_i - \textbf{v}_i \Delta t \right) \\ \nonumber
    &&- \left[ \Delta \tilde{\textbf{p}}_{ij} \left( \bar{\textbf{b}^r_j} \right) + \frac{\partial \Delta \bar{\textbf{p}}_{ij}}{\partial \textbf{b}^r} \delta \textbf{b}^r_j \right] 
    \\
    \textbf{r}_{\textbf{v}_j} &=& \textbf{v}_j - \left[ \tilde{\textbf{v}}_j \left( \bar{\textbf{b}^r_j} \right) + \frac{\partial \bar{\textbf{v}}_j}{\partial \textbf{b}^r} \delta \textbf{b}^r_j \right]
    \\
    {\Vert \textbf{r}_{\Delta \textbf{b}^r_{ij}}\Vert}^2 &=& {\Vert \textbf{b}^r_j - \textbf{b}^r_i \Vert}^2
\end{eqnarray}
, while $\textbf{R}_i$, $\textbf{p}_i$, $\textbf{v}_i$, $\textbf{b}_i^r$ are the orientation, position, velocity, and radar velocity bias at state $i$. 
By dividing orientation state into roll, pitch, and yaw, $\textbf{R}_i$ changes into $\textbf{R}(\gamma_i)\textbf{R}(\beta_i)\textbf{R}(\alpha_i)$ when the $\alpha$, $\beta$, $\gamma$ are roll, pitch, yaw of $\textbf{R}_i$ and $\textbf{R}()$ is $\textbf{Exp} \left(  {{ \left[ \alpha ~ 0 ~ 0 \right] }^\top }_\times \right)$, $\textbf{Exp} \left( { { \left[ 0 ~ \beta ~ 0 \right] }^\top }_\times \right)$, $\textbf{Exp} \left( { { \left[ 0 ~ 0 ~ \gamma \right] }^\top \textbf{}_\times }\right)$ for roll, pitch, yaw each. 
$\textbf{Exp}()$ and $\textbf{Log}()$ an exponential mapping and logarithm mapping in Lie algebra, while $\textbf{v}_\times$ is a skew matrix of the vector $\textbf{v}$. 
As $\textbf{r}_{\textbf{v}_j}$ and ${\Vert \textbf{r}_{\Delta \textbf{b}^r_{ij}}\Vert}^2$ are not including $\textbf{R}_i$, only $\textbf{r}_{\Delta \textbf{p}_{ij}}$ changes:  
\begin{eqnarray}
    \nonumber
    \textbf{r}_{\Delta \textbf{p}_{ij}} &=& 
    {\textbf{R} \left( \alpha_i \right)}^\top
    {\textbf{R} \left( \beta_i \right)}^\top
    {\textbf{R} \left( \gamma_i \right)}^\top
    \left( \textbf{p}_j - \textbf{p}_i - \textbf{v}_i \Delta t \right) \\
    &&- \left[ \Delta \tilde{\textbf{p}}_{ij} \left( \bar{\textbf{b}^r_j} \right) + \frac{\partial \Delta \bar{\textbf{p}}_{ij}}{\partial \textbf{b}^r} \delta \textbf{b}^r_j \right]
\end{eqnarray}
Similarly, orientation state $\textbf{R}_i$ is changed into $\alpha_i$, $\beta_i$, $\gamma_i$ state. 

For the Jacobian, perturbation for $\gamma_i$ is only performed as the roll and pitch values are not optimized.
To simplify the notion using perturbation, $\delta {\gamma}_i^{\textbf{vec}} = {\left[ 0 ~ 0 ~ \delta \gamma_i \right]}^\top$
Using the first-order approximation of the exponential map, the Jacobian of $\textbf{r}_{\Delta \textbf{p}_{ij}}$ about $\gamma_i$ is derivated as follows by starting with changing $\gamma_i$ in $\textbf{r}_{\Delta \textbf{p}_{ij}}$ into $\gamma_i + \delta \gamma_i$:

\footnotesize
\vspace{-4mm}
\begin{eqnarray}
    \nonumber
    \textbf{r}_{\Delta \textbf{p}_{ij}} ' &=& 
    {\textbf{R} \left( \alpha_i \right)}^\top
    {\textbf{R} \left( \beta_i \right)}^\top
    {\textbf{R} \left( \gamma_i + \delta {\gamma}_i^{\textbf{vec}} \right)}^\top
    \left( \textbf{p}_j - \textbf{p}_i - \textbf{v}_i \Delta t \right) \\ 
    &&- \left[ \Delta \tilde{\textbf{p}}_{ij} \left( \bar{\textbf{b}^r_j} \right) + \frac{\partial \Delta \bar{\textbf{p}}_{ij}}{\partial \textbf{b}^r} \delta \textbf{b}^r_j \right]
    \\ \nonumber
    &=&
    {\textbf{R} \left( \alpha_i \right)}^\top
    {\textbf{R} \left( \beta_i \right)}^\top
    \left( \textbf{I} - {\delta {\gamma}_i^{\textbf{vec}}}_\times \right)
    {\textbf{R} \left( \gamma_i \right)}^\top
    \left( \textbf{p}_j - \textbf{p}_i - \textbf{v}_i \Delta t \right) \\ 
    &&- \left[ \Delta \tilde{\textbf{p}}_{ij} \left( \bar{\textbf{b}^r_j} \right) + \frac{\partial \Delta \bar{\textbf{p}}_{ij}}{\partial \textbf{b}^r} \delta \textbf{b}^r_j \right]
    \\ \nonumber
    &=& 
    {\textbf{R} \left( \alpha_i \right)}^\top
    {\textbf{R} \left( \beta_i \right)}^\top
    {\left(
    {\textbf{R} \left( \gamma_i \right)}^\top
    \left( \textbf{p}_j - \textbf{p}_i - \textbf{v}_i \Delta t \right) \right)}_\times
    \delta {\gamma}_i^{\textbf{vec}}
    \\ 
    &&+ ~ \textbf{r}_{\Delta \textbf{p}_{ij}}
\end{eqnarray}
\normalsize

Furthermore, as the $\delta {\gamma}_i^{\textbf{vec}} = {\left[ 0 ~ 0 ~ \delta \gamma_i \right]}^\top$, 
\begin{eqnarray}
\label{eq:blank}
\text{let} ~~ \frac{\partial \textbf{r}_{\Delta \textbf{p}_{ij}}}{\partial \gamma_i} =
    \left[ 
    \begin{array}{ccc}
    a & b & c \\
    d & e & f \\
    g & h & i
    \end{array}
    \right], ~~ \text{then} ~~
    \frac{\partial \textbf{r}_{\Delta \textbf{p}_{ij}}}{\partial \gamma_i} =
    \left[
    \begin{array}{c}
    c \\ f \\ i
    \end{array}
    \right]
\end{eqnarray}
\small
(Third column of 
${\textbf{R} \left( \alpha_i \right)}^\top {\textbf{R} \left( \beta_i \right)}^\top {\left( {\textbf{R} \left( \gamma_i \right)}^\top \left( \textbf{p}_j - \textbf{p}_i - \textbf{v}_i \Delta t \right) \right)}_\times$
)
\normalsize

Based on the derivation of $\frac{\partial \textbf{r}_{\Delta \textbf{p}_{ij}}}{\partial \gamma_i}$, the Jacobian of the $\alpha_i$, $\beta_i$,$\gamma_i$ is as \eqref{eq:jacobian} while most of the matrix is filled with zero due to the 4-DoF optimization and blank area is same as \eqref{eq:blank}, which is not intuitive to write with its components due to its formula. 
\begin{eqnarray}
    \label{eq:jacobian}
    \left[
    \begin{array}{ccc | ccc}
    \frac{\partial \textbf{r}_{\Delta \textbf{p}_{ij}}}{\delta \alpha_i} & 
    \frac{\partial \textbf{r}_{\Delta \textbf{p}_{ij}}}{\delta \beta_i} & 
    \frac{\partial \textbf{r}_{\Delta \textbf{p}_{ij}}}{\delta \gamma_i} & 
    \frac{\partial \textbf{r}_{\Delta \textbf{p}_{ij}}}{\delta \alpha_j} & 
    \frac{\partial \textbf{r}_{\Delta \textbf{p}_{ij}}}{\delta \beta_j} & 
    \frac{\partial \textbf{r}_{\Delta \textbf{p}_{ij}}}{\delta \gamma_j} 
    \\
    \frac{\partial \textbf{r}_{\textbf{v}_j}}{\delta \alpha_i} & 
    \frac{\partial \textbf{r}_{\textbf{v}_j}}{\delta \beta_i} & 
    \frac{\partial \textbf{r}_{\textbf{v}_j}}{\delta \gamma_i} & 
    \frac{\partial \textbf{r}_{\textbf{v}_j}}{\delta \alpha_j} & 
    \frac{\partial \textbf{r}_{\textbf{v}_j}}{\delta \beta_j} & 
    \frac{\partial \textbf{r}_{\textbf{v}_j}}{\delta \gamma_j} 
    \\
    \frac{\partial \textbf{r}_{\Delta \textbf{b}^r_{ij}}}{\delta \alpha_i} & 
    \frac{\partial \textbf{r}_{\Delta \textbf{b}^r_{ij}}}{\delta \beta_i} & 
    \frac{\partial \textbf{r}_{\Delta \textbf{b}^r_{ij}}}{\delta \gamma_i} & 
    \frac{\partial \textbf{r}_{\Delta \textbf{b}^r_{ij}}}{\delta \alpha_j} & 
    \frac{\partial \textbf{r}_{\Delta \textbf{b}^r_{ij}}}{\delta \beta_j} & 
    \frac{\partial \textbf{r}_{\Delta \textbf{b}^r_{ij}}}{\delta \gamma_j} 
    \\
    \end{array}
    \right]
    \\ \nonumber
    =
    \left[
    \begin{array}{ccc | ccc}
    \textbf{0}_{3\times1} &
    \textbf{0}_{3\times1} &
     - &
    \textbf{0}_{3\times1} &
    \textbf{0}_{3\times1} &
    \textbf{0}_{3\times1} 
    \\
    \textbf{0}_{3\times1} &
    \textbf{0}_{3\times1} &
    \textbf{0}_{3\times1} &
    \textbf{0}_{3\times1} &
    \textbf{0}_{3\times1} &
    \textbf{0}_{3\times1}
    \\
    \textbf{0}_{3\times1} &
    \textbf{0}_{3\times1} &
    \textbf{0}_{3\times1} &
    \textbf{0}_{3\times1} &
    \textbf{0}_{3\times1} &
    \textbf{0}_{3\times1} 
    \end{array}
    \right]
\end{eqnarray}
\eqref{eq:jacobian} is the altered orientation part of the residual Jacobian from the original 6-DoF radar factor~\cite{park20213d}. 
% Remaining parts of the jacobian for the other states ($\textbf{p}_i$, $\textbf{p}_j$, $\textbf{v}_i$, $\textbf{v}_j$, $\textbf{b}_i^r$, $\textbf{p}_j^r$) are same as \cite{park20213d} as those states are not changed in the procedure of reducing roll and pitch optimization from the original radar factor. 

\section{Rolling Contact Leg Odometry Factor}
\label{sec:leg_factor}

\begin{figure}[!t]
    \centering
    \includegraphics[width=\columnwidth]{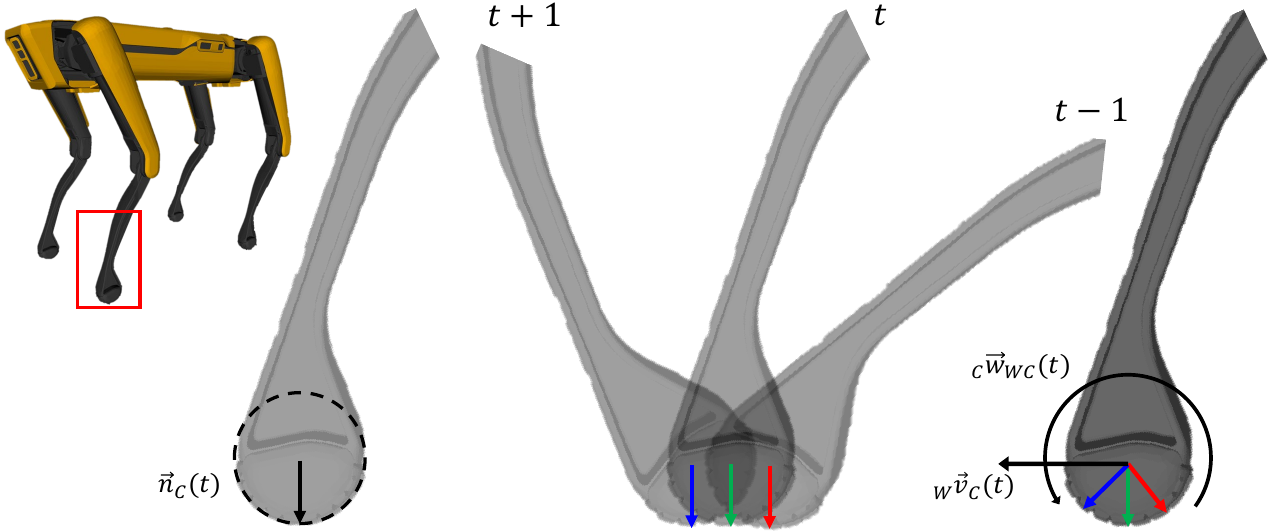}
    \caption{Overview of the rolling contact on legged robots with round feet. Rolling motion on the contact frame occurs as the robot moves. }
\vspace{-5mm}
    \label{fig:leg}
\end{figure}

This section is divided into two subsections. Firstly, we include velocity estimation using kinematic information and rolling contact assumption. Then we apply the preintegrated leg odometry factor introduced from \cite{wisth2022vilens} to the estimated velocity. Unlike the previous radar section, the derivation is more concentrated between ``frames'' not ``time'', $i$ that marks the state $i$ is omitted in the following derivation. 

\subsection{Kinematic Velocity Estimation with Rolling Contact}\label{vel}

Prior to the derivation of robot base velocity, the relationship between the body-contact frame and forward kinematic functions is defined as follows:
\begin{eqnarray}
    \label{eq:fkr}
    \textbf{R}_{\textbf{BC}}&=&\textbf{f}_\text{R} \left( \boldsymbol{\alpha} \right)
    \\
    \label{eq:fkt}
    {}_{\textbf{B}}\textbf{p}_{\textbf{C}}&=&\textbf{f}_{\text{p}} \left( \boldsymbol{\alpha} \right)
\end{eqnarray}
where the $\boldsymbol{\alpha}$ is the measured joint angle and $\textbf{f}_{\text{R}}$, $\textbf{f}_{\text{p}}$ are forward kinematic functions about orientation and translation between robot body frame and contact frame each. 
Both \eqref{eq:fkr} and \eqref{eq:fkt} are differentiated with the Jacobians of each forward kinematic function as follows. 
\begin{eqnarray}
    \dot{\textbf{R}}_{\textbf{BC}} &=& 
    \textbf{R}_{\textbf{BC}}\  {}_{\textbf{C}}^{} \bm{\omega}^{\wedge}_{\textbf{BC}} ~=~
    \textbf{J}_\text{R} \left( \boldsymbol{\alpha} ,  \dot{\boldsymbol{\alpha}} \right) 
    \\
    {}_{\textbf{B}}\dot{\textbf{p}}_{\textbf{C}} &=&
    {}_{\textbf{B}}\textbf{v}_{\textbf{C}} ~=~
    \textbf{J}_\text{p} \left( \boldsymbol{\alpha} ,  \dot{\boldsymbol{\alpha}} \right) 
\end{eqnarray}

Robot base velocity derivation begins from the relationship between the contact and world frames. 
\begin{eqnarray}
    \label{eq:orientation}
    \textbf{R}_{\textbf{WC}} &=&
    \textbf{R}_{\textbf{WB}} \  \textbf{R}_{\textbf{BC}} 
    \\
    \label{eq:position}
    {}_{\textbf{W}}\textbf{p}_{\textbf{C}} &=&
    {}_{\textbf{W}}\textbf{p}_{\textbf{B}} + 
    \textbf{R}_{\textbf{WB}} \  {}_{\textbf{B}}\textbf{p}_{\textbf{C}} 
\end{eqnarray}

Most of the previous works that calculated robot velocity based on the contact kinematics started with the assumption that $\textbf{R}_{\textbf{WC}}$ and ${}_{\textbf{W}} \textbf{p}_{\textbf{C}}$ are zero matrix or zero vector as the contact frame is fixed on the ground~\cite{wisth2022vilens}. 
Following the rolling contact assumption as \figref{fig:leg}, we included the movement of the contact frame with respect to the world frame in the derivation process. 
By differentiating \eqref{eq:orientation}, the skew matrix of the robot base frame's angular velocity with respect to the world frame is derived as follows:

\small
\vspace{-4mm}
\begin{eqnarray}
    \dot{\textbf{R}}_{\textbf{WC}} &=&
    \dot{\textbf{R}}_{\textbf{WB}} \ \textbf{R}_{\textbf{BC}} +
    \textbf{R}_{\textbf{WB}} \ \dot{\textbf{R}}_{\textbf{BC}} 
    \\ \nonumber &=&
    \textbf{R}_{\textbf{WB}} \  {}_{\textbf{B}}^{} \bm{\omega}^{\wedge}_{\textbf{WB}} \  \textbf{R}_{\textbf{BC}} + 
    \textbf{R}_{\textbf{WB}} \  \textbf{R}_{\textbf{BC}} \  {}_{\textbf{C}}^{} \bm{\omega}^{\wedge}_{\textbf{BC}}
    \\ \nonumber &=&
    \textbf{R}_{\textbf{WC}} \ {}_{\textbf{C}}^{} \bm{\omega}^{\wedge}_{\textbf{WC}} = 
    \textbf{R}_{\textbf{WB}} \  \textbf{R}_{\textbf{BC}} \  {}_{\textbf{C}}^{} \bm{\omega}^{\wedge}_{\textbf{WC}}
    \\
    \label{eq:eq8}
    \textbf{R}_{\textbf{BC}} \  {}_{\textbf{C}}^{} \bm{\omega}^{\wedge}_{\textbf{WC}} &=&
    {}_{\textbf{B}}^{} \bm{\omega}^{\wedge}_{\textbf{WB}} \  \textbf{R}_{\textbf{BC}} + 
    \textbf{R}_{\textbf{BC}} \  {}_{\textbf{C}}^{} \bm{\omega}^{\wedge}_{\textbf{BC}}
    \\
    \label{eq:eq9}
    {}_{\textbf{B}}^{} \bm{\omega}^{\wedge}_{\textbf{WB}} &=&
    \textbf{R}_{\textbf{BC}} \ {}_{\textbf{C}}^{} \bm{\omega}^{\wedge}_{\textbf{WC}} \  \textbf{R}_{\textbf{CB}} - 
    \textbf{R}_{\textbf{BC}} \ {}_{\textbf{C}}^{} \bm{\omega}^{\wedge}_{\textbf{BC}} \  \textbf{R}_{\textbf{CB}} 
\end{eqnarray}
\normalsize

As the rolling motion of the contact frame follows the movement of \figref{fig:leg}, the skew matrix of the contact frame's angular velocity with respect to the world frame is derived as follows by exploiting the orientation of the contact frame:
\begin{eqnarray}
    \label{eq:rolling}
    \textbf{R}_{\textbf{WC}}^{\text{t}} &=& 
    \textbf{R}_{\textbf{WC}}^{\text{t}-1} \ 
    \textbf{Exp}\left( {}_{\textbf{C}}^{} \bm{\omega}^{\wedge}_{\textbf{WC}} \left( \text{t} \right) \Delta \text{t} \right)
    \\
    \label{eq:motion}
    {}_{\textbf{C}}^{} \bm{\omega}^{\wedge}_{\textbf{WC}} \left( \text{t} \right) &=&
    \cfrac{1}{\Delta \text{t}} \  \textbf{Log} 
    \left( \textbf{R}_{\textbf{CW}}^{\text{t}-1} \ 
    \textbf{R}_{\textbf{WC}}^{\text{t}} \right)
\end{eqnarray}

\noindent while the orientation of the contact frame is calculated with \ac{IMU} orientation and forward kinematic function. 
Substituting the ${}_{\textbf{C}}^{} \bm{\omega}^{\wedge}_{\textbf{WC}}$ and rotation matrixes in \eqref{eq:eq9} with forward kinematics functions, ${}_{\textbf{B}}^{} \bm{\omega}^{\wedge}_{\textbf{WB}}$ is derived as follow:
\begin{eqnarray}
\small
    \therefore {}_{\textbf{B}}^{} \bm{\omega}^{\wedge}_{\textbf{WB}} &=& \ 
    \textbf{f}_\text{R} \left( \boldsymbol{\alpha}  \right)
    \cfrac{1}{\Delta \text{t}} \  \textbf{Log} 
    \left( \textbf{R}_{\textbf{CW}}^{\text{t}-1} \ \textbf{R}_{\textbf{WC}}^{\text{t}} \right)
    \left( \textbf{f}_\text{R} \left( \boldsymbol{\alpha}  \right) \right)^{\intercal}
    \\ \nonumber 
    &&- \textbf{J}_\text{R} \left( \boldsymbol{\alpha} ,  \dot{\boldsymbol{\alpha}} \right) 
    \left( \textbf{f}_\text{R} \left( \boldsymbol{\alpha} \right) \right)^{\intercal}
\end{eqnarray}
\normalsize

Linear velocity ${}_{\textbf{W}} \textbf{v}_{\textbf{B}}$ is achieved by differentiating the \eqref{eq:position}. 
\begin{eqnarray}
    {}_{\textbf{W}}\dot{\textbf{p}}_{\textbf{C}} &=&
    {}_{\textbf{W}}\dot{\textbf{p}}_{\textbf{B}} + 
    \dot{\textbf{R}}_{\textbf{WB}} \  {}_{\textbf{B}}\textbf{p}_{\textbf{C}} +
    \textbf{R}_{\textbf{WB}} \  {}_{\textbf{B}}\dot{\textbf{p}}_{\textbf{C}} 
    \\ 
    \label{eq:eq14}
    {}_{\textbf{W}}\textbf{v}_{\textbf{B}} &=& 
    - \textbf{R}_{\textbf{WB}} \  {}_{\textbf{B}}^{} \bm{\omega}_{\textbf{WB}}^{\wedge} \  {}_{\textbf{B}}\textbf{p}_{\textbf{C}}
    - \textbf{R}_{\textbf{WB}} \  {}_{\textbf{B}}^{} \dot{\textbf{p}}_{\textbf{C}} 
    + {}_{\textbf{W}}\dot{\textbf{p}}_{\textbf{C}}
    \\ \nonumber &=&
    - \textbf{R}_{\textbf{WB}} \  {}_{\textbf{B}}^{} \bm{\omega}_{\textbf{WB}}^{\wedge} \ 
    \textbf{f}_{\text{p}} \left( \boldsymbol{\alpha} \right)
    - \textbf{R}_{\textbf{WB}} \  \textbf{J}_\text{p} 
    \left( \boldsymbol{\alpha} ,  \dot{\boldsymbol{\alpha}} \right) 
    + {}_{\textbf{W}}\dot{\textbf{p}}_{\textbf{C}} 
\end{eqnarray}

Due to the rolling contact assumption, the linear velocity of the contact frame with respect to the world frame is calculated using the normal vector from ground to contact frame as \figref{fig:leg}: 
${}_{\textbf{W}}\dot{\textbf{p}}_{\textbf{C}}=
\textbf{n}_{\text{C}} \times {}_{\textbf{C}}^{} \bm{\omega}^{\text{ }}_{\textbf{WC}}=
- {}_{\textbf{C}}^{} \bm{\omega}^{\wedge}_{\textbf{WC}} \  \textbf{n}_{\text{C}}$. 
Finally, 6-\ac{DoF} velocity of the robot base in world frame is derived :
\begin{eqnarray}
    {}_{\textbf{B}}^{} \bm{\omega}^{\wedge}_{\textbf{WB}} &=& \ 
    \textbf{f}_\text{R} \left( \boldsymbol{\alpha}  \right)
    \cfrac{1}{\Delta \text{t}} \  \textbf{Log} 
    \left( \textbf{R}_{\textbf{CW}}^{\text{t}-1} \ \textbf{R}_{\textbf{WC}}^{\text{t}} \right)
    \left( \textbf{f}_\text{R} \left( \boldsymbol{\alpha}  \right) \right)^{\intercal}
    \\ \nonumber 
    &&- \textbf{J}_\text{R} \left( \boldsymbol{\alpha} ,  \dot{\boldsymbol{\alpha}} \right) 
    \left( \textbf{f}_\text{R} \left( \boldsymbol{\alpha} \right) \right)^{\intercal}
    \\
    \label{eq:leg_vel}
    {}_{\textbf{W}}\textbf{v}_{\textbf{B}} &=& 
    - \textbf{R}_{\textbf{WB}} \  {}_{\textbf{B}}^{} \bm{\omega}_{\textbf{WB}}^{\wedge} \ 
    \textbf{f}_{\text{p}} \left( \boldsymbol{\alpha} \right)
    - \textbf{R}_{\textbf{WB}} \  \textbf{J}_\text{p} 
    \left( \boldsymbol{\alpha} ,  \dot{\boldsymbol{\alpha}} \right)  
    \\ \nonumber
    &&- {}_{\textbf{C}}^{} \bm{\omega}^{\wedge}_{\textbf{WC}} \  \textbf{n}_{\text{C}}
\end{eqnarray}

\subsection{Preintegrated Leg Odometry Factor}\label{leg_factor}

We apply the rolling contact considered velocity measurement to the preintegrated odometry factor. % of \cite{wisth2022vilens}. 

\subsubsection{Velocity Bias}

Though the rolling contact assumption may handle a portion of the bias on robot velocity, contact drift that may occur due to the slippery or deformable terrain still remains. 
In this manner, we tenant the slowly varying bias term $\textbf{b}^{l}$~\cite{wisth2022vilens} to robot body velocity \eqref{eq:leg_vel} as
\begin{eqnarray}
    \Tilde{\textbf{v}} =  \textbf{v} + \textbf{b}^{l} + \boldsymbol{\eta}^{v}
\end{eqnarray}
with the white noise $\boldsymbol{\eta}^{v}$ based on zero-mean Gaussian. 

\subsubsection{Velocity Measurements Preintegration}

As the measured velocity from \ref{vel} is the linear velocity of the robot body with respect to the world frame, position at time $t_j = t_i + \Delta t$ is 
\begin{eqnarray}
\vspace{-4mm}
    \textbf{p}_j &=& \textbf{p}_i + \int^{t_j}_{t_i} \textbf{v} \left( \tau \right) \text{d}\tau
    \nonumber
    \\ \label{eq:integrat_vel}
    &=& \textbf{p}_i + \sum^{j-1}_{k=i} 
    \left[  \left( {\Tilde{\textbf{v}}}_k  - {\textbf{b}^{l}}_k - {\boldsymbol{\eta}^{v}}_k  \right) \Delta t \right]
\end{eqnarray}
Relative measurement and noise separation are as follows:
\begin{eqnarray}
    \nonumber
    \Delta \textbf{p}_{ij} &=&
    \sum^{j-1}_{k=i} 
    \left[  \left( {\Tilde{\textbf{v}}}_k  - {\textbf{b}^{l}}_k - {\boldsymbol{\eta}^{v}}_k  \right) \Delta t \right]
    \\
    &=& \Delta { \Tilde{\textbf{p}}_{ij}} - \delta \textbf{p}_{ij}
    \\
    \Delta { \Tilde{\textbf{p}}_{ij}} &=& 
    \sum^{j-1}_{k=i} 
    \left[  \left( {\Tilde{\textbf{v}}}_k  - {\textbf{b}^{l}}_k \right) \Delta t \right]
    \\
    \delta \textbf{p}_{ij} &=&
    \sum^{j-1}_{k=i} 
    \left[  {\boldsymbol{\eta}^{v}}_k \Delta t \right]
\end{eqnarray}
while $\Delta { \Tilde{\textbf{p}}_{ij}}$ depends on the slowly varying bias $\textbf{b}^l$. 
To avoid the recomputation of $\Delta { \Tilde{\textbf{p}}_{ij}}$ for every input, we exploit the first order approximation as $\textbf{b} = \bar{\textbf{b}} + \delta \textbf{b}$ following~\cite{forster2016manifold}. 
\begin{eqnarray}
    \Delta \Tilde{\textbf{p}}_{ij} \left( \textbf{b}^l \right)
    \simeq
    \Delta \Tilde{\textbf{p}}_{ij} ( \bar{\textbf{b}}^l )
    + \frac{\partial \Tilde{\textbf{p}}_{ij}}{\partial \textbf{b}^l}
    \delta \textbf{b}^l
\end{eqnarray}

\subsubsection{Residual}

The preintegrated velocity factor residual only includes the translation to give more focus on the effect of 4-DoF optimization from section \ref{sec:radar_factor}. 
\begin{eqnarray}
    \textbf{r}_{\Delta \textbf{p}_{ij}} = 
    \textbf{p}_j - \textbf{p}_i
    - \Delta \Tilde{\textbf{p}}_{ij} \left( \textbf{b}^l \right)
\end{eqnarray}

\section{Experiments}
\label{sec:experiment}

\subsection{Experiment Setup}

\begin{figure}[!t]
    \centering
    \includegraphics[width=\columnwidth]{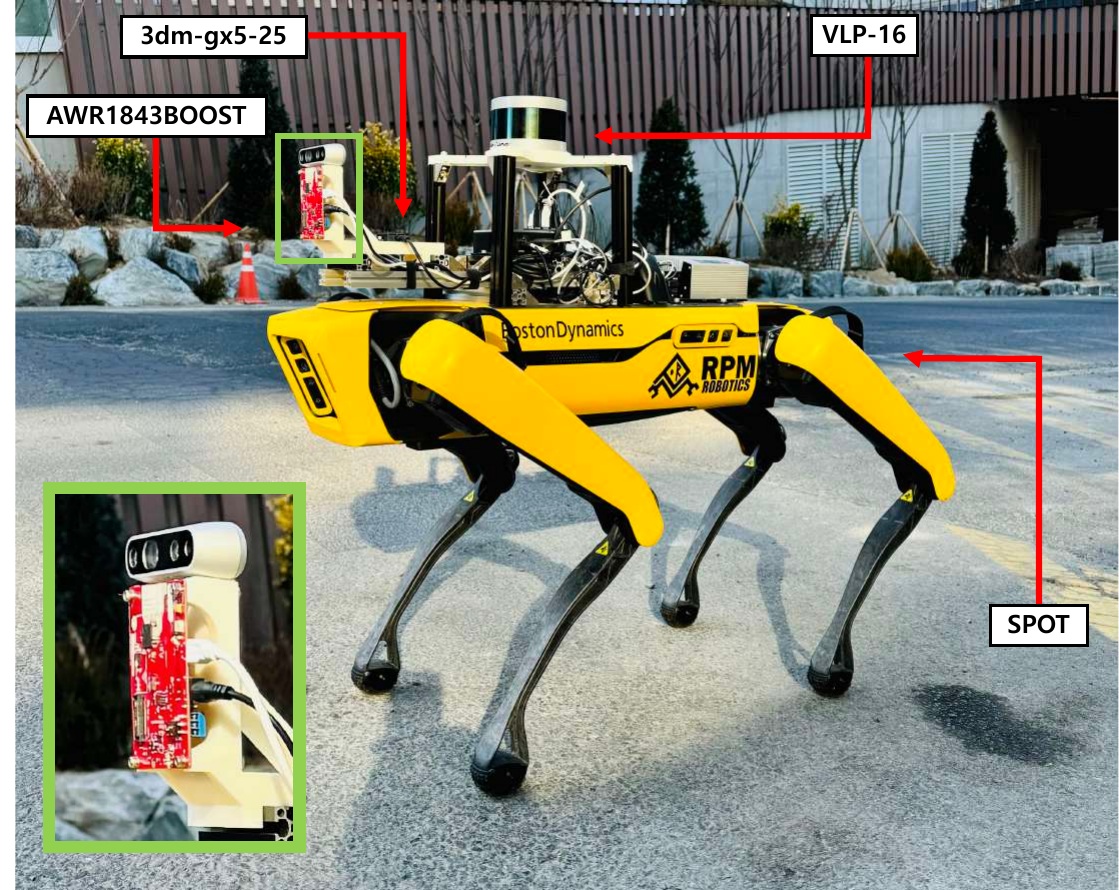}
    \caption{System configuration. LiDAR and camera are attached only for baseline trajectory generation and video recording. }
\vspace{-5mm}
    \label{fig:system}
\end{figure}

\begin{table*}[t]
\centering
\caption{Dataset description}
\label{tab:dataset}
\resizebox{0.9\textwidth}{!}{
\begin{tabular}{c|c|c|c|c|c|c|c|c}
\toprule
\textbf{Sequence}  & \textbf{Environment} & \textbf{Stair} & \textbf{Slope} & \textbf{Narrow Path} & \textbf{Path Length (m)} & \textbf{Elevation Change (m)} & \textbf{Duration (s)} & \textbf{Avg. vel. (m/s)} \\ \midrule
\texttt{Stair} & Outdoor & \cmark & \cmark & \xmark & 253.24 & 15.12 & 304.40 & 0.832 \\ \midrule
\texttt{Under} & Indoor & \xmark & \cmark & \xmark & 227.02 & 3.23 & 258.50 & 0.878 \\ \midrule
\texttt{Narrow} & Outdoor & \xmark & \xmark & \cmark & 134.86 & - & 151.39 & 0.891 \\ \midrule
\texttt{Trail} & Outdoor & \xmark & \cmark & \xmark & 230.10 & 11.37 & 253.36 & 0.908 \\ \midrule
\texttt{Garage} & Hybrid & \xmark & \xmark & \cmark & 138.52 & - & 161.68 & 0.857 \\ \midrule
\texttt{Building} & Hybrid & \cmark & \cmark & \xmark & 252.86 & 8.26 & 291.89 & 0.866
\\ \bottomrule
\end{tabular}}
\vspace{-5mm}
\end{table*}

The system configuration for experiments is shown in \figref{fig:system}. 
We built our sensor system on the SPOT from Boston Dynamics, a legged \ac{UGV} that can provide joint encoder and contact sensor data at \unit{180}{Hz}. 
The AWR1843BOOST radar, which acquires 4D radar pointcloud for \unit{20}{Hz}, and the 3dm-gx5-25 IMU from Microstrain working with \unit{100}{Hz} are exploited. 
Finally, VLP-16 \ac{LiDAR} operating in \unit{10}{Hz} is mounted to generate baseline trajectory. 
% djlee
% exploit이라는 단어가 논문에 엄청 많이 나와서 다른 단어를 써보면 어떨까 싶습니다
% We utilized the sensor system depicted in Figure \ref{fig:system} to collect sensor data for our experiments. Our sensor system is built upon the SPOT quadrupedal robot from Boston Dynamics, which provides joint encoder and contact sensor data at a frequency of 180Hz. Additionally, we utilized the AWR1843BOOST radar, capable of acquiring 4D radar point cloud data at 20Hz, and the 3dm-gx5-25 IMU from MicroStrain operating at 100Hz. Furthermore, we attached a VLP-16 LiDAR operating at 10Hz to generate a baseline trajectory.

\subsubsection{Comparison Targets}\label{comparison}

We compare the proposed method (\texttt{Ours}) against the following five methods. 
\begin{itemize}
    \item \textbf{FAST-LIO2}: FAST-LIO2~\cite{xu2022fast} is utilized to calculate a baseline trajectory (\texttt{LiDAR}). 
    \item \textbf{6-DoF Mono}: Original radar factor of \cite{park20213d} that is working with single radar and 6-DoF optimization (\texttt{Mono}).
    \item \textbf{EKF-RIO}: Best performing open-sourced chip radar odometry method known to us~\cite{doer2020ekf} (\texttt{EKF-RIO}).
    \item \textbf{Leg only}: We only use the preintegrated leg factor to optimize the robot pose for comparison (\texttt{Leg}). 
    \item \textbf{Radar only}: We only use the 4-DoF radar factor to optimize the robot pose for comparison (\texttt{Radar}). 
    \item\textbf{Proposed method}: Our method includes both preintegrated leg factor and 4-DoF radar factor (\texttt{Ours}). 
\end{itemize}

\subsubsection{Dataset}

To evaluate our proposed method, we acquired datasets covering a total of \unit{1.2}{km} for \unit{24}{\min} in various environments such as stairs, slopes, wide and narrow passages, indoors and outdoors, and elevation changes. 
A brief introduction to dataset sequences is as follows:
\begin{itemize}
    \item \texttt{Stair}: \textbf{Stairs that connects the buildings}. Start in front of the building, go up the stairs about \unit{15}{m}, and return to the original position following a slope. 
    \item \texttt{Under}: \textbf{Underground parking lot}. Start from B1, down to B2, and go around the floor. Return to the original location at B1. 
    \item \texttt{Narrow}: \textbf{Walking trails inside university}. Start in front of the building, pass between narrow partitions only a single person can pass through, then walk around the wide space and return to the original position. 
    \item \texttt{Trail}: \textbf{Mountain trail in university}. Start in front of the building and go straight up to the mountain trail; after going up \unit{11}{m}, return to the original location. 
    \item \texttt{Garage}: \textbf{Garage with narrow corridor and outside of the building}. Start from the garage. Go through the corridor and move along the outer wall of the building, returning to the initial location inside the garage. 
    \item \texttt{Building}: \textbf{Inside and outside of the building}. Start moving outside of the building. Move along the slope and get through the underground passage. Climb the stairs about \unit{8}{m} and return to the original position. 
\end{itemize}
\tabref{tab:dataset} includes more detailed information about the dataset.

\subsection{Quantitative Result}

\begin{figure*}[!t]
    \centering
    \includegraphics[width=1\textwidth]{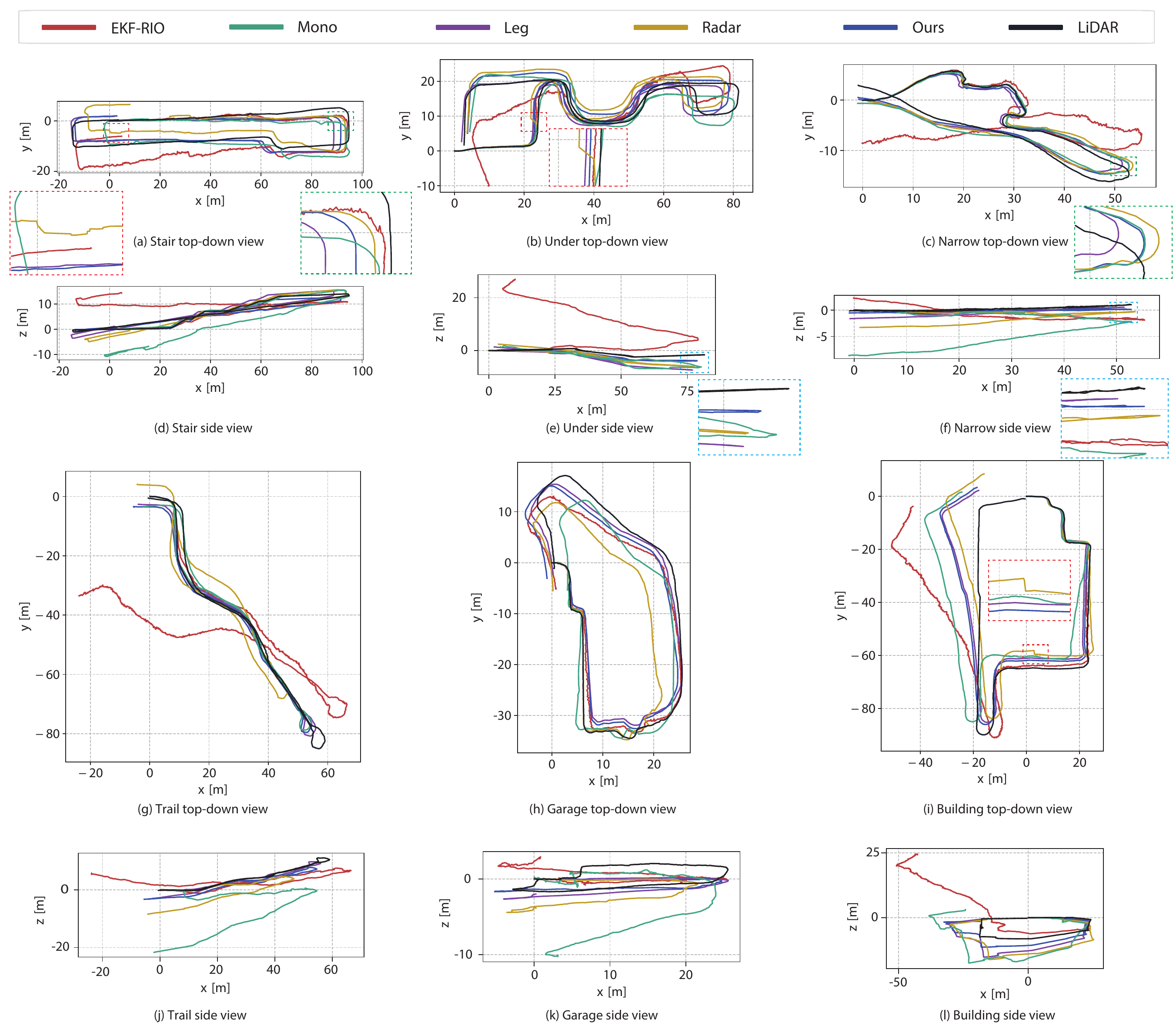}

    \caption{Trajectory result of each sequence's top-down and side views. Color matching of methods is the same for every graph. The red, green, and blue dotted boxes and their enlarged figures are for \ref{study1}, \ref{study2}, and \ref{study3}, respectively. }
    \label{fig:graphs}
\vspace{-6mm}
\end{figure*}

We calculated the \ac{RMSE} of the \ac{ATE} and \ac{RTE} using rpg-trajectory evaluator~\cite{Zhang18iros} as the evaluation metric. 
Additionally, we added \ac{ATE} for the z-axis as a sub-metric to evaluate the performance of the algorithms more focusing on vertical drift. 
ATE$_t$, RTE$_t$, and ATE$_z$ are measured in meters and ATE$_r$, RTE$_r$ are measured in degrees. 
Quantitative results are summarized in \tabref{tab:result} while top-down view and side view plots are included in \figref{fig:graphs}. 
Compared with the \texttt{EKF-RIO} and \texttt{Mono}, the proposing factors provide higher accuracy odometry on most sequences, especially in the z-axis. 
A more detailed analysis is provided in the attached video.

\begin{table}[t]
\centering
\caption{Experimental result. The evaluation was conducted in terms of ATE translation, ATE rotation, RTE translation, RTE rotation, and ATE on the z-axis. The smallest errors from each sequence are marked in \textbf{Bold}. }
\label{tab:result}
\resizebox{0.95\columnwidth}{!}{
\begin{tabular}{c|c|c|c|c|c|c}
\toprule
\multicolumn{2}{c|}{} & \texttt{EKF-RIO} & \texttt{Mono} & \texttt{Leg} & \texttt{Radar} & \texttt{Ours} \\ \midrule
\multirow{5}{*}{\rotatebox{90}{\texttt{Stair}}} & ATE$_t$ & 6.117 & 6.233 & 2.024 & 3.368 & \textbf{1.562} \\
& ATE$_r$ & 4.452 & 8.672 & 3.104 & 3.038 & \textbf{3.022} \\
& RTE$_t$ & 7.110 & 7.473 & 4.167 & 5.037 & \textbf{2.823} \\
& RTE$_r$ & 5.140 & 5.214 & 2.668 & 2.668 & \textbf{2.664} \\
& ATE$_z$ & 5.514 & 3.572 & 0.804 & 1.302 & \textbf{0.717} \\ \midrule
\multirow{5}{*}{\rotatebox{90}{\texttt{Under}}} & ATE$_t$ & 8.945 & 2.343 & 2.470 & 2.903 & \textbf{1.685} \\
& ATE$_r$ & 6.972 & 6.480 & \textbf{1.583} & 1.635 & 1.804 \\
& RTE$_t$ & 9.842 & 4.201 & 3.478 & 3.975 & \textbf{2.417} \\
& RTE$_r$ & 7.370 & 2.814 & 1.801 & 1.801 & \textbf{1.791} \\
& ATE$_z$ & 8.336 & 1.810 & 2.010 & 1.802 & \textbf{0.666} \\ \midrule
\multirow{5}{*}{\rotatebox{90}{\texttt{Narrow}}} & ATE$_t$ & 2.503 & 2.884 & 1.485 & 0.929 & \textbf{0.757} \\
& ATE$_r$ & 7.585 & 6.804 & 2.962 & \textbf{2.708} & 2.745 \\
& RTE$_t$ & 3.950 & 3.720 & 2.483 & 1.799 & \textbf{1.340} \\
& RTE$_r$ & 7.941 & 3.873 & \textbf{2.844} & \textbf{2.844} & 2.845 \\
& ATE$_z$ & 1.264 & 2.765 & 0.928 & 0.339 & \textbf{0.211} \\ \midrule
\multirow{5}{*}{\rotatebox{90}{\texttt{Trail}}} & ATE$_t$ & 10.600 & 6.854 & \textbf{1.943} & 6.374 & 2.586 \\
& ATE$_r$ & 14.891 & 8.167 & \textbf{2.482} & 2.537 & 2.485 \\
& RTE$_t$ & 11.757 & 9.281 & \textbf{3.402} & 9.665 & 4.095 \\
& RTE$_r$ & 16.184 & 4.749 & 3.044 & 3.044 & \textbf{3.039} \\
& ATE$_z$ & 2.333 & 6.314 & \textbf{0.688} & 2.499 & 1.068 \\ \midrule
\multirow{5}{*}{\rotatebox{90}{\texttt{Garage}}} & ATE$_t$ & 2.519 & 3.759 & \textbf{1.305} & 3.108 & 1.494 \\
& ATE$_r$ & 5.201 & 6.454 & 3.968 & \textbf{3.961} & 4.018 \\
& RTE$_t$ & 2.484 & 5.065 & 2.217 & 4.704 & \textbf{2.137} \\
& RTE$_r$ & 4.940 & 3.920 & 2.572 & 2.572 & \textbf{2.569} \\
& ATE$_z$ & 1.850 & 3.006 & \textbf{0.575} & 0.946 & 0.695 \\ \midrule
\multirow{5}{*}{\rotatebox{90}{\texttt{Building}}} & ATE$_t$ & 13.205 & 7.766 & 5.215 & 5.645 & \textbf{4.922} \\
& ATE$_r$ & 16.462 & 10.737 & 8.716 & \textbf{8.596} & 8.781 \\
& RTE$_t$ & 15.361 & 7.832 & 6.148 & 7.005 & \textbf{5.176} \\
& RTE$_r$ & 20.253 & 10.687 & 10.516 & 10.516 & \textbf{10.510} \\
& ATE$_z$ & 7.946 & 3.944 & 2.418 & 2.679 & \textbf{1.167}
\\ \bottomrule
\end{tabular}}
\vspace{-3mm}
\end{table}

\subsection{Radar Drift Reduction by Leg Factor}\label{study1}

Combining leg odometry with a radar sensor reduces odometry drift that may occur due to the occlusion of large dynamic objects(e.g, buses and cars as in \figref{fig:fig_1}). 
In \figref{fig:graphs}(a), \figref{fig:graphs}(b), and \figref{fig:graphs}(i), it can be found that drift has occurred on \texttt{Radar} trajectory which is caused by the cars and pedestrians passing at close range. 
Nevertheless, \texttt{Ours} trajectory showed robustness on drift by tightly fusing radar and leg. 
Furthermore, though the consecutive drift has occurred on \texttt{Trail} sequence as it can be found from \figref{fig:graphs}(g), \texttt{Ours} estimated reasonable odometry, more relying on the information from kinematic sensors. 
By fusing leg odometry factor with radar factor, \texttt{Ours} generated robust odometry to radar velocity drift. 

\subsection{Leg Velocity Scale Handling by Radar Factor}\label{study2}

Though the leg odometry may show robustness and stability in every dataset, due to the contact impact that occurs at every step, inevitable contact drift may not be solved with rolling contact assumption and bias term. 
This leads to the leg velocity scale issue, making \texttt{Leg} trajectory reduced, compared to the \texttt{LiDAR} trajectory. 
However, the radar sensor is free from the scale issue of the leg factor, which can be found in \figref{fig:graphs}(a) and \figref{fig:graphs}(c). 
In this manner, despite the occurrence of velocity drift in radar sensors, it may enhance the odometry calculation being fused with the leg odometry.

\subsection{Accurate z-axis Trajectory Estimation}\label{study3}

As in \tabref{tab:result}, \texttt{Radar} and \texttt{Leg} calculate precise odometry compared with the comparison group even when used alone. 
However, \texttt{Ours} shows more accurate odometry results, especially in consequence of the z-axis, which means that the radar and leg odometry factor may work complementarily to increase the precision of the trajectory. 
Vertical accuracy enhancement by fusing both sensors can be found in \figref{fig:graphs}(e) and \figref{fig:graphs}(f). 
The only sequence that \texttt{Ours} shows a higher error tendency compared with another algorithm is the \texttt{Trail} sequence.
As in \ref{study1}, this is because consecutive drift occurred in the \texttt{Radar} while \texttt{Ours} is well following the \texttt{Leg} trajectory. 
\texttt{Leg} operates complementary to the failed \texttt{Radar}, which can be found on both \figref{fig:graphs}(g) and \figref{fig:graphs}(j). 

\subsection{Robust on Vision Failure}\label{study4}

One of the key characteristics of our radar factor is that it does not exploit the spatial information to measure the robot's velocity. 
Similarly, the leg odometry factor utilizes the contact sensor, joint encoder, and angular velocity of \ac{IMU}, all proprioceptive sensors. 
Because of this, proposed factors are highly robust on vision failure that might occur on camera or \ac{LiDAR}. 
In \figref{fig:graphs}(c), it can be found that the \ac{LiDAR} trajectory is not converging to the starting position, which means that scan-matching drift happened on \texttt{LiDAR} because of the narrow path included in the dataset. 
In contrast, \texttt{Ours}, \texttt{Radar}, \texttt{Leg} show the converging trajectory to the initial position at \figref{fig:graphs}(c) and \texttt{Ours} generate even accurate results on the z-axis as \figref{fig:graphs}(f).  
These findings suggest that the sensor fusion between radar and leg kinematic sensors is highly robust in vision failure environments.

\begin{table}[t]
\centering
\caption{Experimental result for radar factor ablation. Error-values refer to ATE$_z$ (ATE on the z-axis). Results in the first and fourth columns are the same with \texttt{Mono} and \texttt{Radar} in Table. II. }
\label{tab:ablation1}
\resizebox{\columnwidth}{!}{
\begin{tabular}{c|c|c|c|c}
\toprule
RANSAC & \multicolumn{2}{c|}{\textbf{w/o z-axis outlier reduction}} & \multicolumn{2}{c}{\textbf{with z-axis outlier reduction}}  \\ \midrule
Optimization & \hspace*{3mm} \textbf{6-DoF} \hspace*{3mm} & \hspace*{3mm} \textbf{4-DoF} \hspace*{3mm} & \hspace*{3mm} \textbf{6-DoF} \hspace*{3mm}  & \hspace*{3mm}  \textbf{4-DoF} \hspace*{3mm}  \\ \midrule
\texttt{Stair} & 3.572& 3.306 & 2.384 & 1.302 \\ \midrule
\texttt{Under} & 1.810 & 2.030 & 1.559 & 1.802 \\ \midrule
\texttt{Narrow} & 2.765 & 2.293 & 1.375 & 0.339 \\ \midrule
\texttt{Trail} & 6.314 & 6.271 & 3.684 & 2.499 \\ \midrule
\texttt{Garage} & 3.006 & 2.313 & 1.520 & 0.946 \\ \midrule
\texttt{Building} & 3.944 & 4.229 & 3.063 & 2.679 
\\ \bottomrule
\end{tabular}}
    \centering
    \includegraphics[width=\columnwidth]{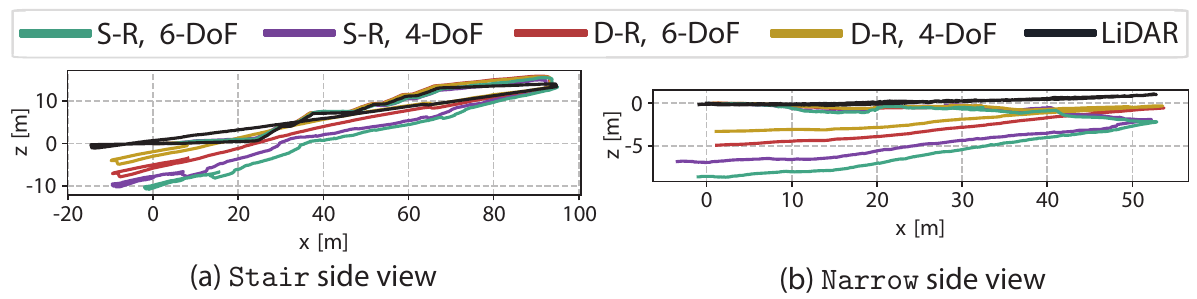}

    \captionof{figure}{Effects of decoupled RANSAC and 4-DoF optimization to the estimated trajectory. S-R and D-R refer to single or decoupled RANSAC, while 4-DoF and 6-DoF refer to the optimization. }
    \label{fig:ablation1}

\end{table}

\begin{table}[t]
\centering
\caption{Experimental result for leg factor ablation. Results on the second column are the same with \texttt{Leg} in Table. II. }
\label{tab:ablation2}
\resizebox{0.8\columnwidth}{!}{
\begin{tabular}{c|c|c|c}
\toprule
\multicolumn{2}{c|}{} & \textbf{w/o rolling contact} & \textbf{with rolling contact} \\ \midrule
\multirow{2}{*}{\texttt{Stair}} & ATE$_t$ & 12.998 & \textbf{2.024} \\ 
& ATE$_z$ & 1.610 & \textbf{0.804} \\ \midrule
\multirow{2}{*}{\texttt{Under}} & ATE$_t$ & 3.261 & \textbf{2.470} \\
& ATE$_z$ & \textbf{0.671} & 2.010 \\ \midrule
\multirow{2}{*}{\texttt{Narrow}} & ATE$_t$ & 5.953 & \textbf{1.485} \\
& ATE$_z$ & \textbf{0.368} & 0.928 \\ \midrule
\multirow{2}{*}{\texttt{Trail}} & ATE$_t$ & 2.378 & \textbf{1.943} \\
& ATE$_z$ & \textbf{0.611} & 0.688 \\ \midrule
\multirow{2}{*}{\texttt{Garage}} & ATE$_t$ & 3.541 & \textbf{1.305} \\
& ATE$_z$ & 1.094 & \textbf{0.575} \\ \midrule
\multirow{2}{*}{\texttt{Building}} & ATE$_t$ & 27.094 & \textbf{5.215} \\
& ATE$_z$ & \textbf{0.544} & 2.418
\\ \bottomrule
\end{tabular}}

    \centering
    \includegraphics[width=\columnwidth]{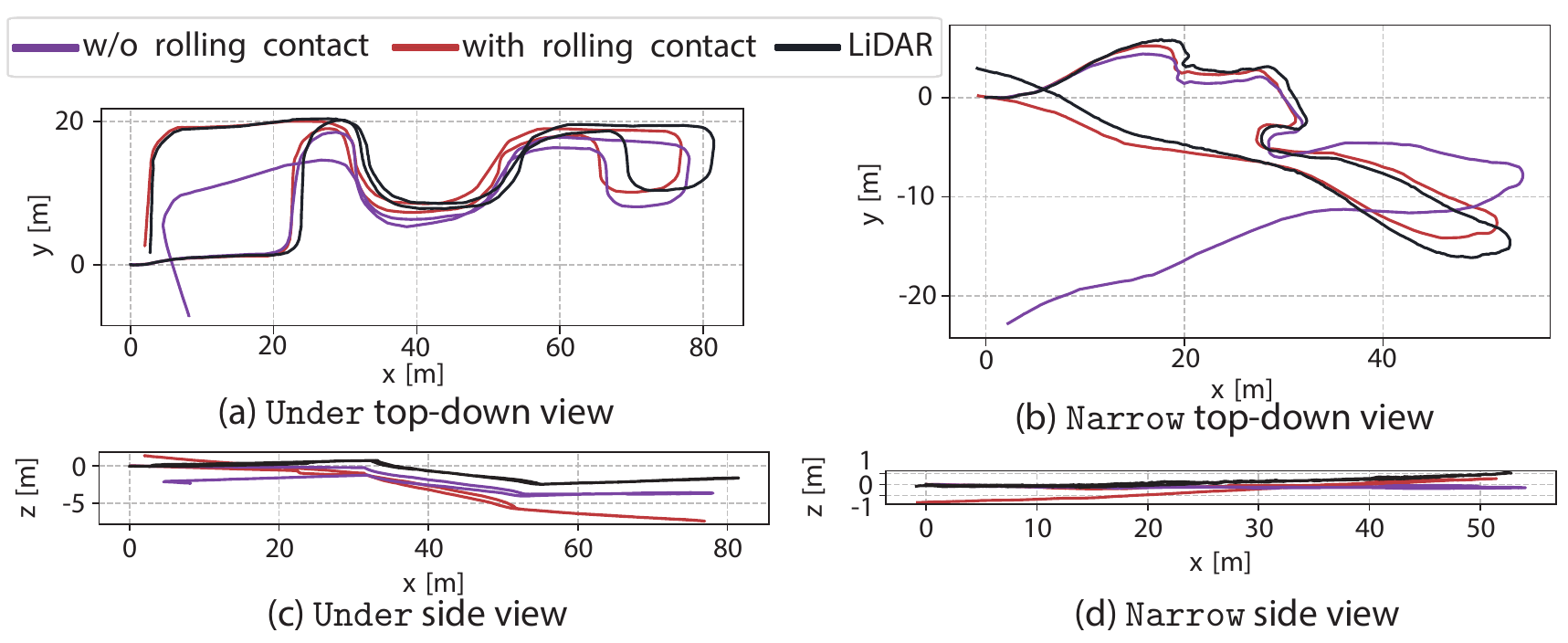}

    \captionof{figure}{Effect of rolling contact to the estimated trajectory. }
    \label{fig:ablation2}

\vspace{-5mm}
\end{table}

\subsection{Effects of Decoupled RANSAC \& 4-DoF Optimization}\label{study5}
Evaluation of the performance of decoupled \ac{RANSAC} and 4-\ac{DoF} optimization in section \ref{sec:radar_factor} are done, and results of the experiment are included in \tabref{tab:ablation1}. 
Outlier reduction using decoupled \ac{RANSAC} effectively increases z-axis precision compared with the original single \ac{RANSAC} algorithm, no matter which optimization scheme is utilized. 
4-\ac{DoF} optimization provides higher precision on the z-axis for most cases, but 6-\ac{DoF} optimization occationally gives better results when the decoupled \ac{RANSAC} is not exploited. 
This concludes that accurate ego-velocity measurement is required to effectively leverage 4-\ac{DoF} optimization, and decoupled \ac{RANSAC} performs that role well. 
Example trajectories about each combination are included in \figref{fig:ablation1}. 

\subsection{Effect of Rolling Contact Assumption}\label{study6}
The effect of rolling contact assumption in section \ref{sec:leg_factor} is done, and experiment results are included in \tabref{tab:ablation2}. 
Though rolling contact assumption shows a subtle contribution to improving z-axis accuracy, it can be confirmed that it effectively improves overall trajectory precision. 
This is because, as found from \figref{fig:leg}, the rolling contact assumption affects the velocity component on the $xy$ plane, not in the vertical direction. 
The significant improvement in horizontal velocity accuracy demonstrates that the rolling contact assumption effectively lowers the burden of leg velocity bias. 
Example trajectories are included in \figref{fig:ablation2}.

\subsection{Computational Cost}

Lastly, we compute computational cost analysis for the full system. 
Since our algorithm leverages sparse radar, contact sensors, and joint encoders, the computational burden is very low. 
The chip radar we exploited has an sampling rate of \unit{20}{Hz}, and the proposed system's computational time was \unit{14.21}{\ms} per sequence on average. 
Time consumption was evaluated on an Intel i7-11700 CPU with \unit{64}{GB} RAM. 
\section{Conclusion}
\label{sec:conclusion}

In this paper, we propose two novel factors to generate more precise and robust odometry for the legged robots. 
By leveraging the additional z-directional noise removal and 4-DoF factor graph optimization scheme, we achieved remarkable performance improvements on radar odometry. 
Furthermore, based on the rolling contact assumption, we lowered the burden of the bias term in velocity estimation and enhanced the preintegrated odometry factor exploiting the velocity. 
We evaluated the performance of proposed factors based on the self-collected dataset from various environments, including stairs, slopes, trails, and dynamic objects.
The proposed method exhibits robustness and consistent performance even when a single sensor fails, as both sensors work in a complementary manner.

\balance
\small
\bibliographystyle{IEEEtranN} %citeauthor
\bibliography{string-short,references}

\end{document}